\def\year{2022}\relax
\documentclass[letterpaper]{article} 
\usepackage{aaai22}  
\usepackage{times}  
\usepackage{helvet}  
\usepackage{courier}  
\usepackage[hyphens]{url}  
\usepackage{graphicx} 
\usepackage{natbib}  
\usepackage{caption} 
\DeclareCaptionStyle{ruled}{labelfont=normalfont,labelsep=colon,strut=off} 
\usepackage{algorithm}
\usepackage{algorithmic}

\usepackage{amssymb}
\usepackage{amsmath}
\usepackage{marvosym}
\usepackage{bm}
\usepackage{ulem}
\usepackage{subfigure}
  
\usepackage{newfloat}
\usepackage{listings}
\floatstyle{ruled}
\newfloat{listing}{tb}{lst}{}
\floatname{listing}{Listing}
\title{Soft Hierarchical Graph Recurrent Networks\\ for Many-Agent Partially Observable Environments}

\author{
    Zhenhui Ye\textsuperscript{\rm 1}
    Xiaohong Jiang\textsuperscript{\rm 2},
    Guanghua Song\textsuperscript{\rm 1*}
    Bowei Yang\textsuperscript{\rm 1}\\
}
\affiliations{
    \textsuperscript{\rm 1}School of Aerospace and Astronautics, Zhejiang University\\
    \textsuperscript{\rm 2} College of Computer Science and Technology, Zhejiang University\\
	\{zhenhuiye, jiangxh, ghsong, boweiy\}@zju.edu.cn
}

\begin{document}

\maketitle

\begin{abstract}
The recent progress in multi-agent deep reinforcement learning(MADRL) makes it more practical in real-world tasks, but its relatively poor scalability and the partially observable constraints raise challenges to its performance and deployment. Based on our intuitive observation that the human society could be regarded as a large-scale partially observable environment, where each individual has the function of communicating with neighbors and remembering its own experience, we propose a novel network structure called hierarchical graph recurrent network(HGRN) for multi-agent cooperation under partial observability. Specifically, we construct the multi-agent system as a graph, use the hierarchical graph attention network(HGAT) to achieve communication between neighboring agents, and exploit GRU to enable agents to record historical information. To encourage exploration and improve robustness, we design a maximum-entropy learning method to learn stochastic policies of a configurable target action entropy. Based on the above technologies, we proposed a value-based MADRL algorithm called Soft-HGRN and its actor-critic variant named SAC-HRGN. Experimental results based on three homogeneous tasks and one heterogeneous environment not only show that our approach achieves clear improvements compared with four baselines, but also demonstrates the interpretability, scalability, and transferability of the proposed model. Ablation studies prove the function and necessity of each component.
\end{abstract}

\section{Introduction}
Human society could achieve efficient communication and collaboration, which is the foundation of general artificial intelligence. In essence, human society could be seen as a large-scale partially observable multi-agent system, in which the automation of many collaborative tasks will bring great efficiency improvements. In recent years, multi-agent deep reinforcement learning(MADRL) has been facilitated to solve real-life problems such as package routing\cite{package_routing1,package_routing2} and unmanned aerial vehicles(UAVs) control\cite{UAV_control}, which are typical of large-scale and partial observability. To solve the environmental instability in the training process of multi-agent systems, MADDPG\cite{MADDPG} introduces the framework of centralized training and decentralized execution(CTDE) and leads to many variants. However, CT brings high computational complexity and poor scalability, and DE makes it difficult for agents to obtain the necessary information for collaboration under partial observability. To handle the information insufficiency in partially observable environments, DIAL\cite{DIAL2016} pioneered the paradigm of communication learning. It performs communication among all agents, which makes it hard to extract valuable information for cooperation. An alternative is to allow agents to communicate to others within a certain range,  R-MADDPG\cite{R-MADDPG} proposes to communicate with a fixed number of agents, which may restrain the range of cooperation or incur the loss of important information that could help cooperation. In this case, to achieve efficient neighboring communication, a scalable and flexible information aggregation mechanism is needed. The recent progress of deep learning and graph learning provides a new idea for MARL, i.e., to regard the multi-agent system as a graph. Under such intuition, DGN\cite{DGN} and HAMA\cite{HAMA} adopt the graph attention network(GAT)\cite{GAT2018} as the communication structure to achieve better scalability and better performance in multi-agent environments. This paper, on the basis of these works, further explores a highly scalable MADRL algorithm for large-scale partially observable environments.

Inspired by the observation that each individual in human society obtains necessary information for collaboration by communicating with the colleges and recalling its own experience, we propose a network structure called \textit{hierarchical graph recurrent network} (HGRN). HGRN regards the multi-agent system as a graph, each agent as a node with its local observation as the node embedding, and requires prior knowledge (such as distance, network connectivity, etc) to connect the nodes. To achieve communication between heterogeneous agents, we adopt \textit{hierarchical graph attention network} (HGAT) \cite{HAMA} with small modifications as the graph convolutional layer, which employs self-attention as the convolutional kernel and extracts valuable information from different groups of neighboring agents respectively. The role of HGAT is very similar to the convolutional layer in CNN, and its output is the node encoding of the local embedding and information aggregated from the one-hop neighbors. Apart from spatial information aggregation, it is helpful to recall valuable information in temporal histories under partial observability. Thus we use \textit{gated recurrent unit} (GRU)\cite{GRU2014} to grant agents the ability to keep valuable memory for a long time. In this way, HRGN could make decisions based on the information aggregated from its neighbors and its memory. 

Apart from the network structure, the exploration strategy is also critical in the POMDP environment. Instead of learning deterministic policies with heuristic exploration methods such as $\epsilon$-greedy as many previous works did, we propose a maximum-entropy method that could learn stochastic policies of a configurable target action entropy. As the need for exploration varies from different scenarios, the optimal target action entropy also varies. Thus our strategy is more interpretable and convenient to find the optimal setting compared with MAAC\cite{MAAC2019} that learns stochastic policies with a fixed temperature parameter. We name the value-based policies trained in this way as Soft-HGRN, and its actor-critic variant as SAC-HGRN.

The main contributions of this paper are as follows: 
\begin{itemize}
	\item[(a)] We introduce the HGRN structure that combines the advantages of HGAT and GRU. It enables the network to obtain spatio-temporal information to handle the partially observable environment.
	\item[(b)] We propose two maximum-entropy MADRL models that introduce a learnable temperature parameter to learn our HGRN-structured policy with a configurable target action entropy, namely Soft-HGRN and SAC-HGRN.
	\item[(c)] We show that our approach outperforms four state-of-art MADRL methods in several homogeneous and heterogeneous environments. Case studies and ablation studies interpret the function and necessity of each component in our methods.
\end{itemize}

\section{Related Works}
To solve the problem of multi-agent cooperation, the simplest and straightforward method is to use single-agent deep reinforcement learning (SADRL) algorithm to train each agent independently. This method belongs to the decentralized training and decentralized execution(DTDE) paradigm and is known as independent learning. However, training multiple policies at the same time may make the environment too unstable for the model to converge. To solve the environmental instability, MADDPG\cite{MADDPG} extends DDPG\cite{DDPG} by learning a centralized critic network with full observation to update the decentralized actor network with partial observability. This paradigm is known as centralized training and decentralized execution(CTDE) and leads to many variants such as MAAC\cite{MAAC2019} and PEDMA\cite{PEDMA}. However, since the input space of centralized critic expands exponentially to the scale of multi-agent system, it is hard to converge in large-scale multi-agent tasks. As a consequence, many large-scale multi-agent cooperative tasks\cite{fuzzyQ}\cite{traffic_light} are still handled by independent learning methods such as DQN\cite{DQN} and A2C\cite{A2C}.

Communication learning aims to learn a communication protocol among agents to enhance cooperation. DIAL\cite{DIAL2016} is the first work that proposes a learnable communication protocol between agents in the partially observable environment. CommNet\cite{CommNet} proposes to use the average of the embedding of all agents as the global communication value. Note that both of them assume that all agents would communicate with each other, which leads to poor scalability. In recent years, the development of graph neural networks(GNNs) brings a scalable and flexible communication structure for MADRL. \textit{Networked agent}\cite{fully_decentralized_networked_agents} constructs the multi-agent system into a graph and transfers the network parameters along the edges; DGN\cite{DGN} proposes to stack two GAT layers to achieve inter-agent communication in a two-hop perception region; HAMA\cite{HAMA} proposes the hierarchical graph attention network(HGAT) structure, which achieves the collaboration among multiple agent groups. Most recently, \cite{niu2021multi} propose a graph attention communication protocol with a scheduler to decide when to communicate and whom to address messages to.

The proposed Soft-HGRN and SAC-HGRN are similar to DGN and HAMA in that it utilizes the graph attention mechanism for inter-agent communication. However, DGN only considers communication among homogeneous agents, and HAMA has not devoted much attention to designing the overall network structure. By contrast, Soft-HGRN improves the HGAT layer to better communicate with heterogeneous agents and properly design the HGAT-based network structure. DRQN\cite{DRQN},  QMIX\cite{QMIX} and  R-MADDPG\cite{R-MADDPG} also use recurrent units to store historical information to address POMDP problems. Our novelty is that the stored historical information is the aggregated information obtained by graph convolution. MAAC\cite{MAAC2019} also uses SAC\cite{SAC} to train stochastic policies and they set a fixed temperature parameter to control the exploration, yet we learn the temperature parameter according to the configurable target action entropy, which shows better interpretability.

\section{Background and Notations}

A partially observable Markov Game(POMG) is a multi-agent extension of Markov decision process(MDP). At each timeslot of a POMG environment with $N$ agent, each agent $i$  obtains a local observation $o_i$ and executes an action $a_i$, then receives a scalar reward $r_i$ from the environment. The objective of reinforcement learning(RL) is to learn an policy $\pi_i(a_i|o_i)$ for agent $i$ that maximizes its discounted reward $\mathbb{E}[R_t]=\mathbb{E}[\Sigma_{t=0}^T \gamma^t r_i^t]$, where $\gamma\in[0,1]$ is a discounting factor.  Our work is based on the framework of POMG with neighboring communication, i.e, each agent could obtain extra information from its neighbors.

Q-learning\cite{q-learning} is a popular method in RL and have been widely used in multi-agent domains\cite{independant_q_learning}. It learns a value function $Q(o,a)$ that estimate the expected return $\mathbb{E}[\Sigma_{t=\tau}^T \gamma^t r_i^t]$ after taking action $a$ under observation $o$, which could be recursively defined as $Q(o_t,a_t)=\mathbb{E}_{a_{t+1}}[r+Q(o_{t+1},a_{t+1})]$. DQN\cite{DQN} is the first work that learns a Q-function with the deep neural network as its approximator, which introduces experience replay\cite{experience_replay} and target network to stabilize the training. At each environmental timeslot $t$, it stores the transition tuple (namely the \textit{experience}) $(o_t, a_t, r_t,o_{t+1})$ into a large sliding window container(namely the \textit{replay buffer}), and resample a mini-batch of experience from the replay buffer in every $\tau$ steps. Then update the Q-function by minimizing the loss function as follows:

\begin{equation}
Q_{loss} = (r_t+\max_{a_{t+1}} Q'(o_{t+1},a_{t+1}) - Q(o_t,a_t))^2
\end{equation}
where $Q'$ is the target network whose parameters are periodically updated by copying the parameters of the learned network $Q$.  Now that an action-value function $Q$  is trained, a optimal policy can be obtained by selecting the action with biggest Q-value: $\pi^*(a|o)=\arg \max_a Q(o,a)$. As the greedy policy is easy to converge to sub-optimal, DQN is trained (and generally executed) with heuristic exploration strategies such as $\epsilon$-greedy.

Graph attention network(GAT)\cite{GAT2018} is a remarkable work in Deep Learning and is regarded as a powerful network structure to calculate the relationships between agents in MADRL. Generally, the agent $i$ in the environment could be represented as a node with its local information $e_i$ as the node embedding.  The connection between nodes can be determined by distance, network connectivity, or other metrics. For convenience, we use $G_i$  to represent the set of agent $i$ and its neighbors. To aggregate valuable information from its neighbors, agent $i$ would calculate its relationship to each neighbor $j\in G_i$ with a bilinear transform\cite{Attention}: 
\begin{equation}
\alpha_{ij} = \frac{\exp((W_K e_j)^T\cdot W_Q e_i)}{\Sigma_{j\in G_i} \exp((W_K e_j)^T\cdot W_Q e_i)}
\end{equation}
then computes the aggregated node embedding as the output of GAT layer: $g_i= \Sigma_{j \in G_i} \alpha_{ij} \cdot W_Ve_j$, where $W_Q,W_K,W_V$ are learnable matrices that transform the node embedding into the query, key, and value vector.

Due to limited space, a table of important notations is presented in Appendix.

\section{Soft Hierarchical Graph Recurrent Networks}
In this section, we introduce our MADRL approach for large-scale partially observable problems, including a novel network structure named HGRN  and two maximum-entropy MADRL methods named Soft-HGRN and SAC-HGRN.
\subsection{Aggregate information from neighbors and histories}
The design of the hierarchical graph recurrent network(HGRN) is inspired by human society, in which each individual can communicate with its (logical or physical) neighbors and recall valuable information from its memory. We construct the multi-agent system into a graph, where each agent in the environment is represented as a node and its local information as the node embedding $e_i$. The node embedding could pass through the edge during the forward propagation. For each agent $i$, it has agent $j\in G_i$ within its communication range, where $G_i$ represents the set of neighboring agents that interconnect to agent $i$. Since we consider heterogeneous settings, there could be $K$ groups of agents in the environment, and we represent the set of agents in group $k$ as $C^k$. There are mainly two challenges in designing a network structure for communication among heterogeneous agents: First, the graph structure will dynamically change over time, which requires each node to have good scalability and robustness to process the neighboring nodes' information; second, the fea ture representation between different groups of agents may vary greatly, which makes it challenging to utilize the features from heterogeneous agents.

We adopt the idea of \textit{hierarchical graph attention network} (HGAT) \cite{HAMA} to execute communication among heterogeneous agents. The key idea of HGAT is to communicate by group, i.e., computes a separate node embedding vector $g_i^k$ for each agent group $k=1,\ldots,K$. Specifically, consider an agent $i$, to extract information from agent group $k$ to form the node embedding $g_i^k$, we first compute the individual relationship between agent $i$ to each of its neighbors $j$ in the group $k$, represented as: 

\begin{equation}
	\alpha^k_{ij} = \frac{\exp((W^k_K e_j)^T\cdot W^1_Q e_i)}{\Sigma_{j\in G_i\cap C^k} \exp((W^k_K e_j)^T\cdot W^1_Q e_i)}
\end{equation}
then we aggregate the information of each neighbor $j$ by:

\begin{equation}
	g_i^k = \Sigma_{j \in G_i\cap C^k} \alpha^k_{ij} \cdot W_V^k e_j
\end{equation}

Different from vanilla HGAT that uses a shared self-attention unit to integrate the group-level embeddings, we concatenate the embeddings of each group and process it with a linear transform:

\begin{equation}
	g_i = W(g_i^1|...|g_i^K)
\end{equation}
where $(\cdot|\cdot)$ denotes the concatenation operation. We perform this modification for two purposes: first, the characteristics of different groups of agents can be quite different, and using shared parameters to process their embedding may lead to training instability; the second is to achieve the unity of the network structure, i.e, when all agents are homogeneous, our version of HGAT is equivalent to GAT. A figure that illustrates the detailed structure of our modified HGAT can be found in Appendix.

After using HGAT to realize the spatial information aggregation, we hope that the agent could keep valuable temporal information in memory, which requires the agent to have the ability to judge and select critical information in its history. To this end, we apply the \textit{gated recurrent unit}(GRU) to process the node embedding $g_i$:
\begin{equation}
h_{t} = GRU(g_{t}|h_{t-1})
\end{equation}
where $h_{t-1}$ is the hidden state stored in the GRU, which records valuable information in historical node embeddings.

\begin{figure}[t]
	\centering
	\includegraphics[width=0.9\columnwidth]{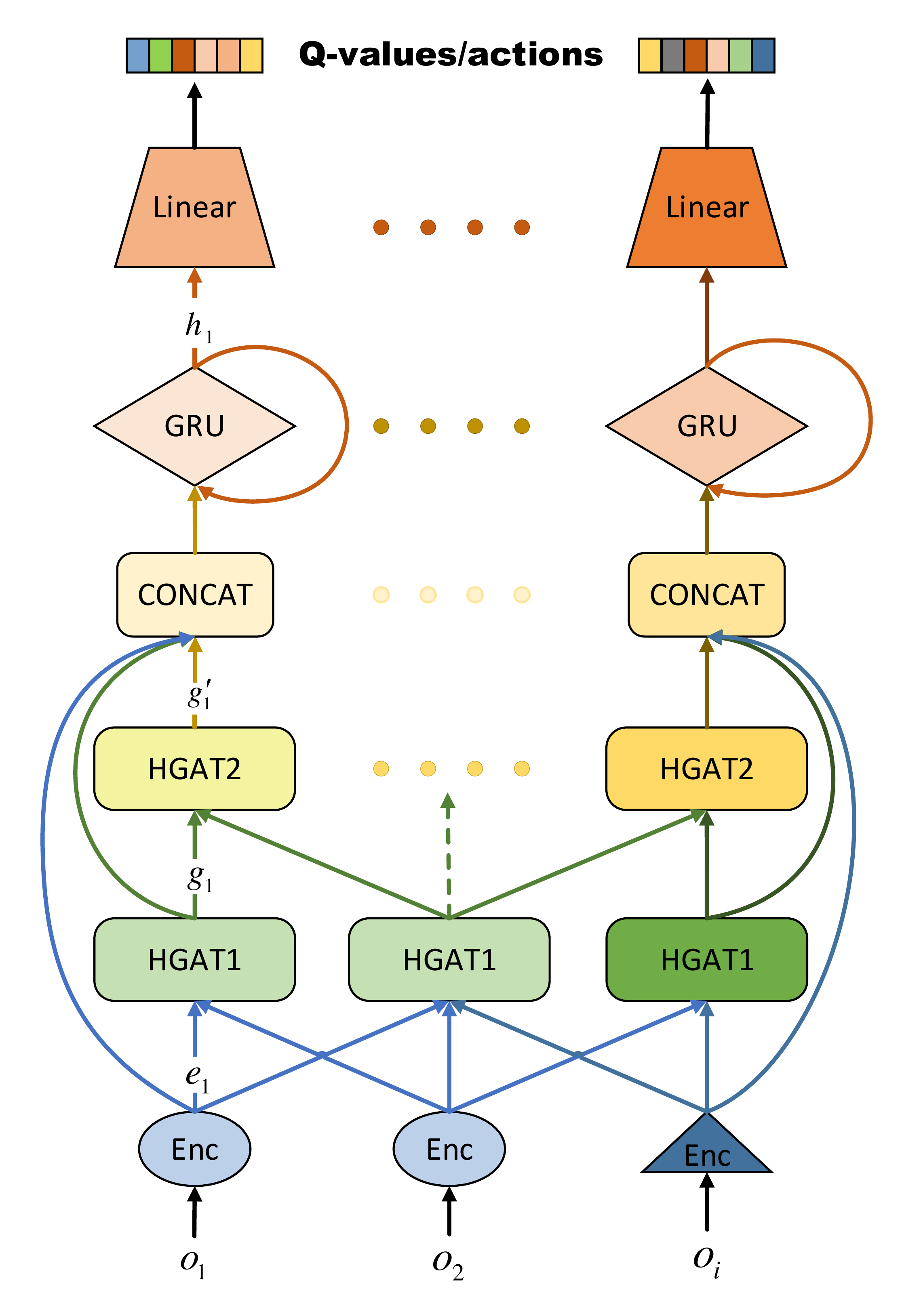} 
	\caption{The overall structure of HGRN.}.
	\label{fig:hgrn-network}
\end{figure}

The overall network structure of our proposed HGRN is shown in Figure \ref{fig:hgrn-network}. For scalability and sample efficiency, agents in the same group share the same parameters. To reduce the computational complexity, HGRN only takes observations as the input and outputs the Q-value for each possible action. We use MLP encoders to map the raw observation of heterogeneous agents to the same dimension, and then use the encoding as the node embedding in HGAT. We perform a network architecture search to design the HGAT-based communication structure, details can be found in Appendix. Compared with HAMA\cite{HAMA} that only uses one HGAT layer, our proposed HGRN stacks two HGAT layers and therefore has a two-hop perceptive field. Skip connections\cite{ResNet} are also created over the two HGAT layers to achieve better performance and faster convergence.  Behind the GRU, a linear transform is applied to infer the Q-values for all actions.

For clarity, we use $\mathcal{O}_t^i$ to represent the set of observation of agent $i$ and its neighbors at time $t$, i.e., $\mathcal{O}_t^i=\{o_t^j|\forall j\in G_t^i\}$ . Hence the HGRN-structured Q-function could be represented as $Q^i:\mathcal{O}_t^i\rightarrow \mathbb{R}^\mathcal{A}$, where $\mathcal{A}$ is the dimension of the action space.

\subsection{Learn stochastic policies with configurable action entropy}
The deterministic policy, such as DQN and DGN, is easy to fall into a local optimum, hence $\epsilon$-greedy is often used in the training phase and execution phase to encourage exploration. However, the exploration induced by $\epsilon$-greedy is completely random and may incur performance degradation during execution. Our intuition is that stochastic policy can be used to replace the $\epsilon$-greedy to help exploration, which allows the model to learn when and how to explore. To this end, we adopt soft Q-learning\cite{soft_Q-learning} to train an energy-based stochastic model. Specifically, we get the probability of the action by using softmax to process the Q value:
\begin{equation}
\begin{split}
\pi(a_t|\mathcal{O}_t)= \frac{\exp\Big(\frac{1}{\alpha}\cdot Q(\mathcal{O}_t,a_t)\Big)}{\Sigma_{a_t}\exp\Big(\frac{1}{\alpha}\cdot Q(\mathcal{O}_t,a_t)\Big)}\\
=\exp\Big(\frac{Q(\mathcal{O}_t,a_t)}{\alpha}-\log\Sigma_{a_t} \exp\Big(\frac{Q(\mathcal{O}_{t},a_{t})}{\alpha}\Big)\Big)
\end{split}
\end{equation}
where $\alpha$ is a temperature parameter to control the level of exploration of the model, which is proportional to the diversity of the policy's output actions. The larger the $\alpha$, the more balanced the action probability distribution of the policy. To learn such an energy-based policy, the value function $V$ is redefined as:
\begin{equation}
	V(\mathcal{O}_t)=\alpha\cdot\log\Sigma_{a_t} \exp\Big(\frac{Q(\mathcal{O}_{t},a_{t})}{\alpha}\Big)
\end{equation}
Then the Q-function is updated by minimizing the mean squared TD-error:
\begin{equation}
	Q\_loss = \frac{1}{S}\Sigma\Big(r_t + V(\mathcal{O}_{t+1})-Q(\mathcal{O}_t,a_t)\Big)^2
\end{equation}
where $S$ is the size of mini-batch, $y_t=r_t + V(\mathcal{O}_{t+1})$ is the learning target of $Q(\mathcal{O}_t,a_t)$.

In general, the action entropy is widely used to measure the degree of exploration of a policy, which is defined as the information entropy of the policy's action probability:
\begin{equation}
	\mathbb{E}[\mathcal{H}_\pi(\mathcal{O})]=\mathbb{E}[- \Sigma_a\pi(a|\mathcal{O})\cdot \log\pi(a|\mathcal{O})]
\end{equation}

\begin{figure}[t]
	\centering
	\includegraphics[width=0.9\columnwidth]{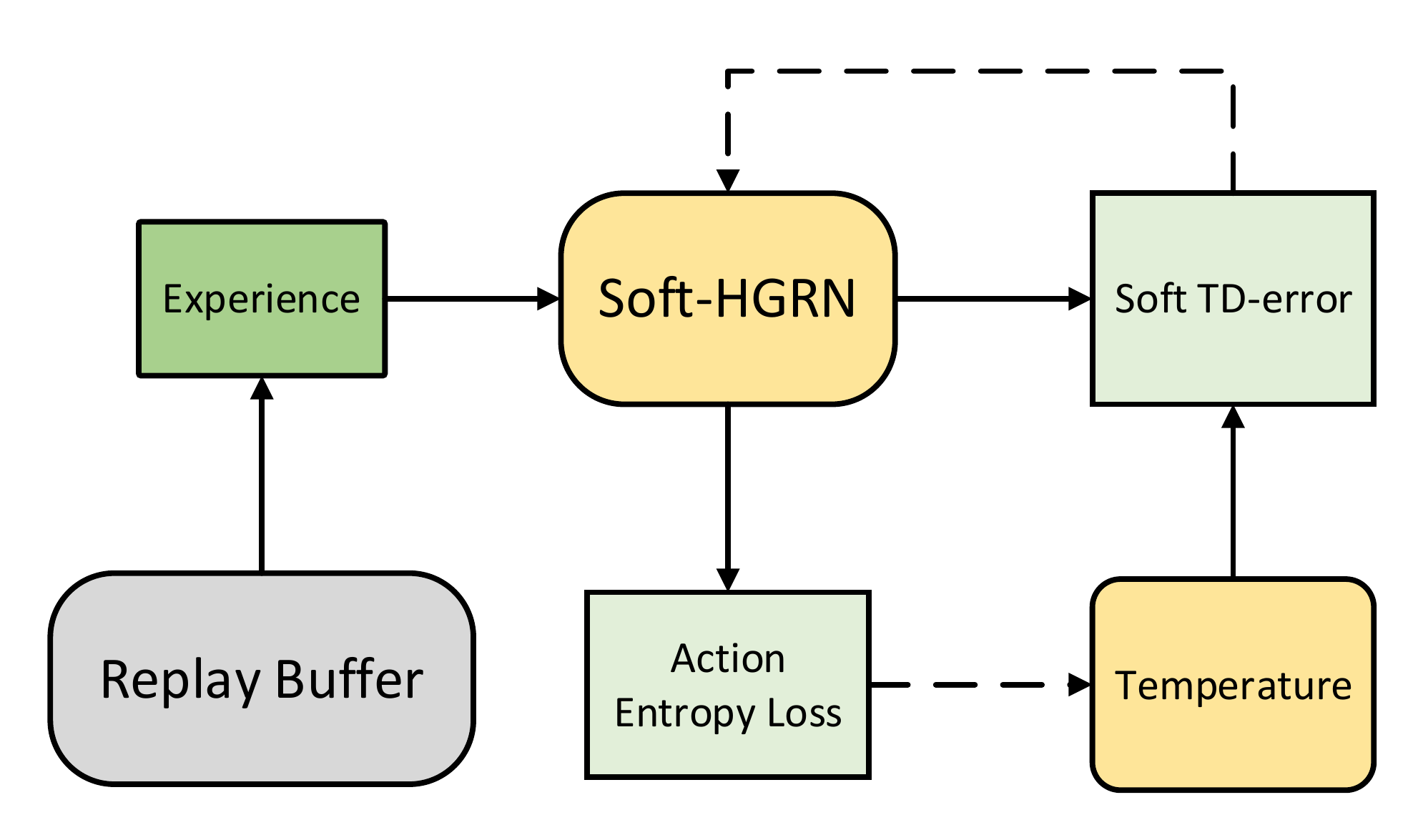} 
	\caption{The flowchart of learning process of soft-HGRN. Solid lines denote the forward propagation to calculate the loss, and the dashed lines denote backward propagation.}.
	\label{fig:soft-hgrn-flowchart}
\end{figure}

As is mentioned above, the action entropy is controlled by the temperature hyper-parameter $\alpha$ . However, the effect of $\alpha$ on the action entropy varies with different network structures and environments. By contrast, setting the action entropy as the goal of the learned model is more intuitive and interpretable. Therefore, we set a target action entropy $\mathcal{H}_{target}$, and then adaptively adjust the temperature parameter $\alpha$ to approximate the goal of $\mathbb{E}[\mathcal{H}_\pi (\mathcal{O})]\rightarrow \mathcal{H}_{target}$. Specifically, we learn the $\alpha$ with gradient descent by:
\begin{equation}
\nabla_{\alpha}=f\big(\mathcal{H}_{target}-\mathbb{E}[\mathcal{H}_{\pi}(\mathcal{O})] \big)
\end{equation}
where $f$ is a function that meets the condition of $f(x)\cdot x \leq 0$. Also note that $\mathcal{H}_{target}$ is the target entropy value for the policy to approximate and we generally represent it as  $\mathcal{H}_{target}=p_{\alpha}\cdot\max\mathcal{H}_\pi  $, where $\max \mathcal{H}_\pi$ is determined by the action space of the policy $\pi$, and $p_\alpha\in[0,1]$ is a new hyper-parameter to be tuned. We name the approach that learns the HGRN-based stochastic model with configurable action entropy as Soft-HGRN. To be intuitive, a flowchart of the learning process of Soft-HGRN is shown in Figure \ref{fig:soft-hgrn-flowchart}.

In some complicated scenarios, decoupling the task of value estimation and action selection can improve the performance of the model. Adapted from SAC\cite{SAC}, we design an actor-critic styled variant named SAC-HGRN, in which the value function $V$ is defined as:
\begin{equation}
	\begin{split}
	V(\mathcal{O}_t)= \mathbb{E}_{a_t\sim \pi(a_t|\mathcal{O}_t)} [Q(\mathcal{O}_t,a_{t}) - \alpha\log\pi(a_{t}|\mathcal{O}_t)]
	\end{split}
\end{equation}
and the actor network is updated with policy gradient with a maximum entropy regularization:
\begin{equation}
	\begin{split}
	\nabla \pi = \frac{1}{S}\Sigma_S\nabla \log \pi(a_t|\mathcal{O}_t) \cdot\Big(Q(\mathcal{O}_t,a_t)- \alpha\log\pi(a_t|\mathcal{O}_t)\Big)
	\end{split}
\end{equation}
Due to limited space, more details about Soft-HGRN and SAC-DRGN, as well as the pseudocode can be found in Appendix.

\section{Experiments}
We apply our methods in four many-agent partially observable environments and mainly aim to answer the following questions:
\begin{itemize}
	\item \textbf{Q1:} How does Soft-HGRN/SAC-HGRN perform in the homogeneous and heterogeneous tasks, compared with other baselines from the state-of-the-art? 
	\item \textbf{Q2:} How does the HGRN structure extract necessary features for effective cooperation?
	\item \textbf{Q3:} How does the learnable temperature of soft Q-learning work to control the action entropy and improve the performance of the policy?
\end{itemize}
Several additional experiments including parameter analysis, interpretability study, and all learning curves could be found in Appendix.

\subsection{Environments}
The tested environments include three homogeneous(\textit{UAV-MBS}, \textit{Surviving}, and \textit{Pursuit}) and one heterogeneous scenario (\textit{Cooperative Treasure Collection}).  An overview of the environments is as follows, cf. Appendix for detailed settings and screenshots.

\begin{itemize}
	\item \textbf{UAV-MBS}\cite{ye_2021} is a cooperative task with 20 UAVs served as mobile base stations to fly around a target region to provide communication services to the randomly distributed ground users.
	\item \textbf{Surviving}\cite{DGN} is a moral dilemma consisting of 100 agents and a limited amount of food. The agents should cooperate to explore a big map to discover the randomly refreshed food but then compete with each other to occupy the food as soon as possible to prevent starvation. This environment is more complicated than UAV-MBS since the explored food will be consumed and be randomly regenerated in another position.
	\item \textbf{Pursuit}\cite{MAgent} consists of 25 learnable predators and 50 pre-trained prey. The predators need to cooperate with nearby teammates to form several types of closure to lock their prey.
	\item \textbf{Cooperative Treasure Collection}\cite{MAAC2019} is a heterogeneous environment with three types of agents, 20 \textit{hunter} units which can collect the red and blue treasure, and 10 \textit{red/blue bank} units which can store the treasure of the corresponding color from the hunter.
\end{itemize}

\subsection{Baseline Methods}
	We compared our approach with four MADRL baselines, including DQN, CommNet, MAAC, and DGN. A more detailed discussion on the characteristics of each algorithm can be found in Appendix. 

\subsection{Implementation Details}
We conduct experiments on the four environments with our methods and four baselines. Each model is updated until convergence. We use Adam\cite{Adam2015} as the optimizer of the network, and use SGD with a same learning rate to update the learnable temperature parameter $\alpha$. To make a fair comparison, the communication structure of DGN and MAAC is implemented as two stacked GAT with skip-connection, which is similar to HGRN as shown in Figure \ref{fig:hgrn-network}. More hyper-parameter settings and the network structure of all models can be found in Appendix.

\subsection{Performance Evaluation and Ablation Studies}
We first compare the performance of Soft-HGRN and SAC-HGRN with other baselines to answer \textbf{Q1}. 

For the homogeneous environments, we train the model until convergence and use the mean episodic reward over 2,000 testing episodes as our evaluation metric. We remove each component (HGAT, GRU, and Stochastic policy) from our approach and test its performance to clarify the function and necessity of each component. We conduct three repeated runs for each case and calculate the averaged results along with their standard deviations, as shown in Table \ref{table:result_rewards}. Note that HGAT is equivalent to GAT in homogeneous tasks, hence DGN is equivalent to (Soft-HGRN - G - S) in the table. It can be seen that our approach outperforms other baselines by a significant margin, and removing any component would cause performance degradation.  All learning curves can be found in Appendix.

\begin{table}[!t]
	\small
	\renewcommand{\arraystretch}{1.2}
	\centering
	\begin{tabular}{|c||c|c|c|}
		\hline
		\textbf{Algorithm}& \textbf{UAV-MBS} &\textbf{Surviving}& \textbf{Pursuit}\\
		\hline
		DQN & $2314\pm177$ & $-7453\pm74$ &$5007\pm 150$ \\
		CommNet & $2377\pm 116$ & $-7511\pm82$ &$5222\pm 60$ \\
		MAAC & $3435\pm 88$& $\uuline{-10\pm46}$ &$5927\pm 91$  \\
		DGN & $2808\pm 109$ &$-286\pm67$& $4552\pm50$ \\
		\hline
		Soft-HGRN & \uline{$4051\pm 75$}&$\uline{194\pm41}$ & $6840\pm20$\\
		Soft-HGRN - G &$3751\pm59$&$-118\pm90$ & $\uline{7138\pm96}$\\
		Soft-HGRN - R &$2935\pm 134$&$-26\pm61$ & $5649\pm 166$\\
		Soft-HGRN - S &\uuline{$3925\pm 46$}&$-89\pm35$ & $5570\pm30$\\
		\hline
		SAC- HGRN & \bm{$4072\pm77$}&$\bm{325\pm48}$ & $\uuline{7033\pm55}$\\
		SAC-HGRN - G & $3580\pm75$&$-36\pm63$ & $\bm{7183\pm77}$\\
		SAC-HGRN - R& $3052\pm22$&$141\pm13$ & $5797\pm36$\\
		SAC-HGRN - S& $3723\pm82$&$-376\pm27$ & $-76\pm1$\\
		\hline
	\end{tabular}
	\caption{The episodic reward of different methods in three homogeneous tasks. '-G' means disabling HGAT-based communication among agents, '-R' means removing GRU-based memory unit from the policy model, '-S' denotes training a deterministic policy instead of a stochastic policy. The best, second and third models are indicated by bold, single underline, and double underline respectively.}
	\label{table:result_rewards}
\end{table}

We also notice that the necessity of each component varies in different environments, possibly due to the different nature of each task. Firstly, removing GRU in UAV-MBS and Pursuit will significantly drop the performance while it is not that obvious in Surviving. This finding denotes that the temporal memory is of necessity in UAV-MBS and Pursuit, possibly since the position of ground users in UAV-MBS is fixed during an episode, and predators in Pursuit should construct consistent closure to lock the prey; while in Surviving the food will be quickly consumed and regenerated in another position. Secondly, it can be seen that HGAT-based communication is ultra important in Surviving, possibly because the dynamic food distribution requires cooperative exploration among the agents. A notable finding is that disabling the communication leads to the best performance in Pursuit, our insight is that the key factor in this task is to form a stable closure with neighboring predators, which only requires information of very near neighbors that in the local observation range, and information from far away may distract the agent. Thirdly, it can be seen that the performance of the stochastic policies trained with maximum-entropy learning outperforms the corresponding deterministic policies, proving its capability to substitute $\epsilon$-greedy to provide a more intelligent exploration. Finally, it can be seen that SAC-HGRN shows better performance over Soft-HGRN in Table \ref{table:result_rewards}. Although our environments are made of discrete grids, SAC-HGRN can easily be applied to continuous space. 

\begin{figure}[t]
	\centering
	\includegraphics[width=0.9\columnwidth]{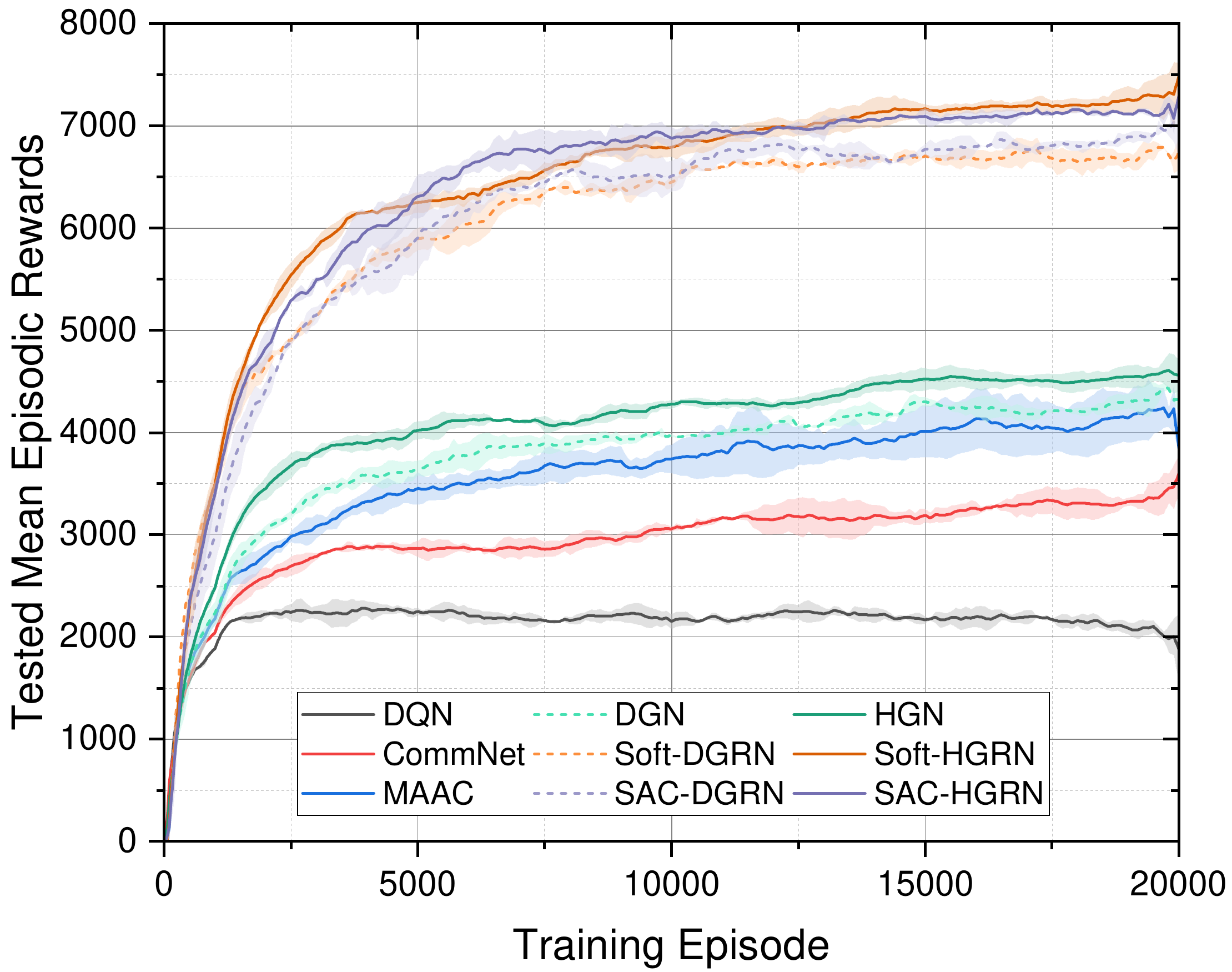} 
	\caption{Learning curves of various algorithms in CTC. }.
	\label{fig:learning_curves_ctc}
\end{figure}

In the heterogeneous environment CTC, we mainly focus on the impact of HGAT's communication between different types of agents. To this end, we also test the DGN with HGAT(namely HGN), and Soft-HGRN/SAC-HGRN without heterogeneous communication(namely Soft-DGRN and SAC-DGRN).  The learning curves of all tested models are shown in Figure \ref{fig:learning_curves_ctc}, and error bars denote the standard derivation over three runs. It can be seen that every HGAT-based model (the solid line) performs better than its GAT-based variant (the dashed line),  which demonstrates the effectiveness of our designed hierarchical communication structure.

\subsection{Case Studies about Interpretability}
In this section, we perform a series of case studies to investigate the principle behind the performance of our approach. 

\subsubsection{How HGRN extracts valuable features?}
\begin{figure}[t]
	\centering
	\subfigure[A screenshot of the intuitive example. ]{
		\includegraphics[width=0.85\columnwidth]{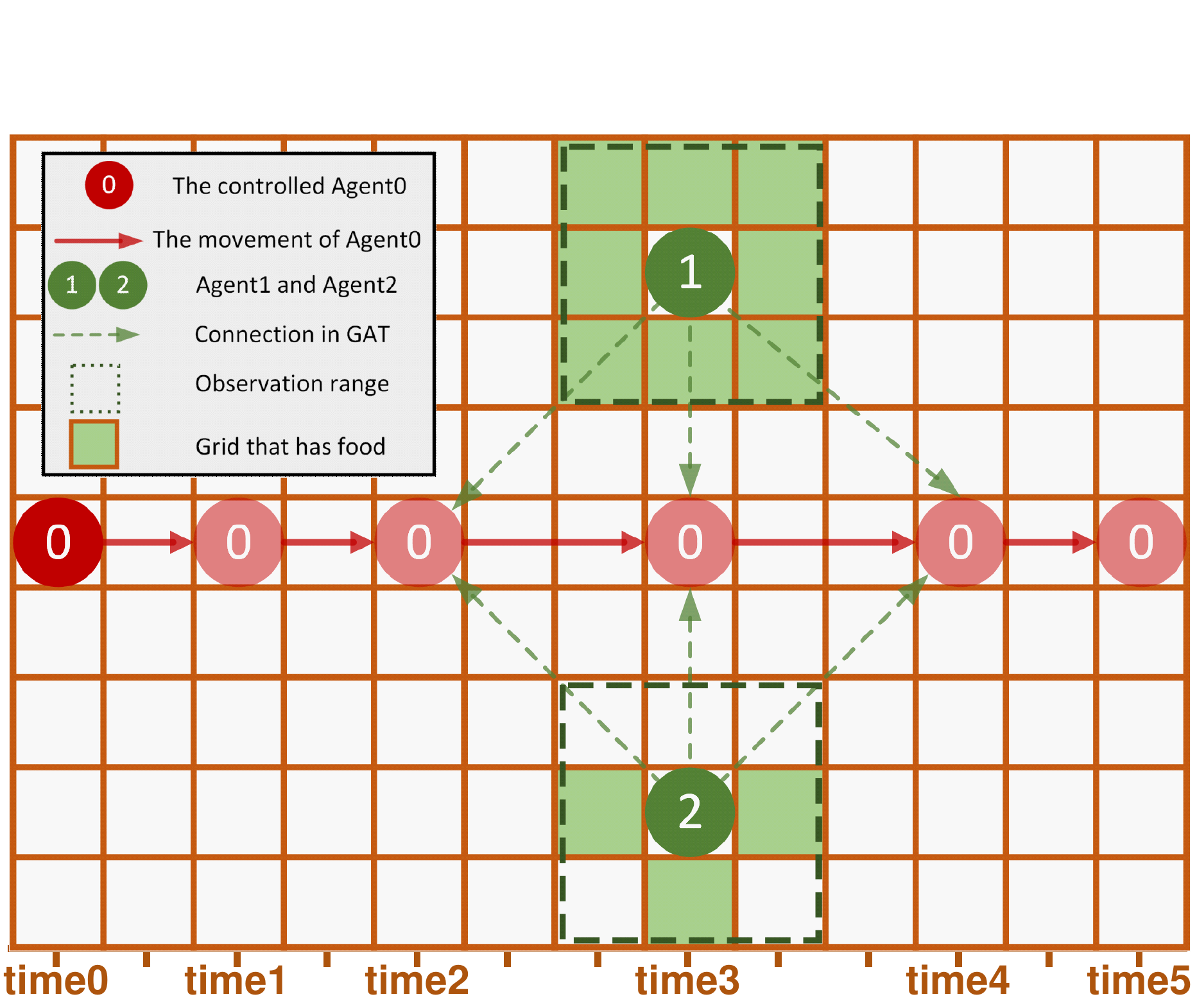} 
		\label{fig:example}
	}
	\quad
	\subfigure[The graph attention weights of agent 0.]{
		\includegraphics[width=0.9\columnwidth]{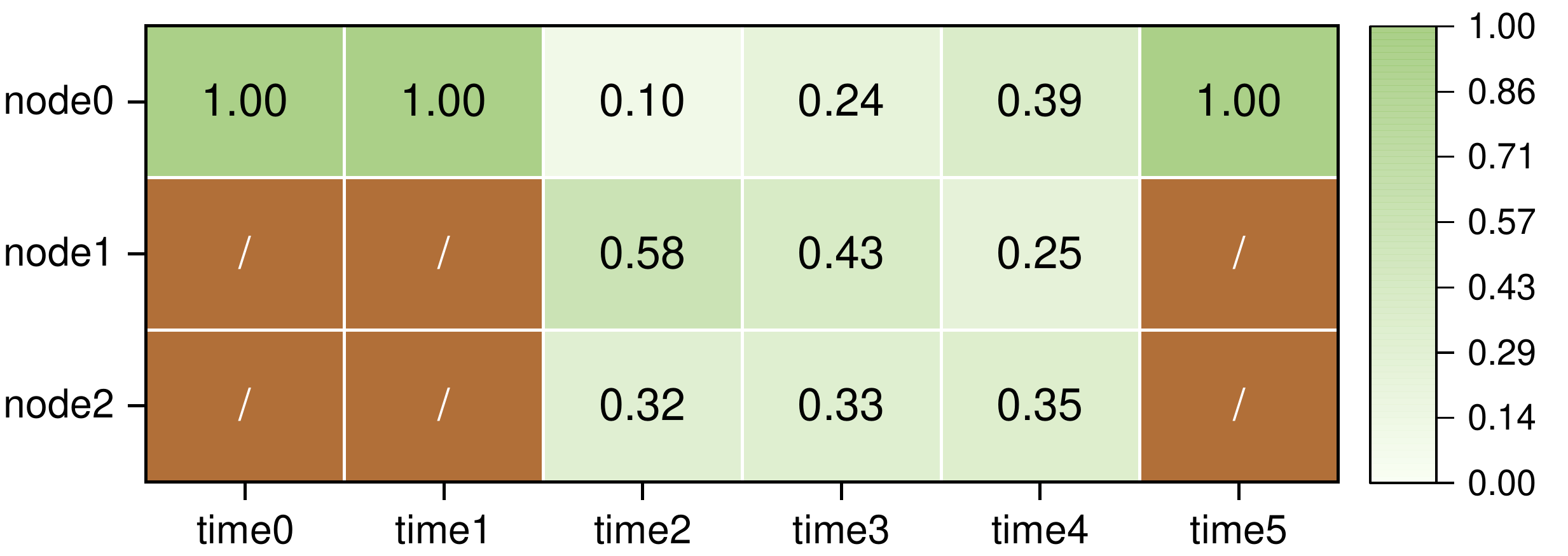} 
		\label{fig:attention_time}
	}
	\quad
	\subfigure[The mean value of GRU's reset gate.]{	
		\includegraphics[width=0.8\columnwidth]{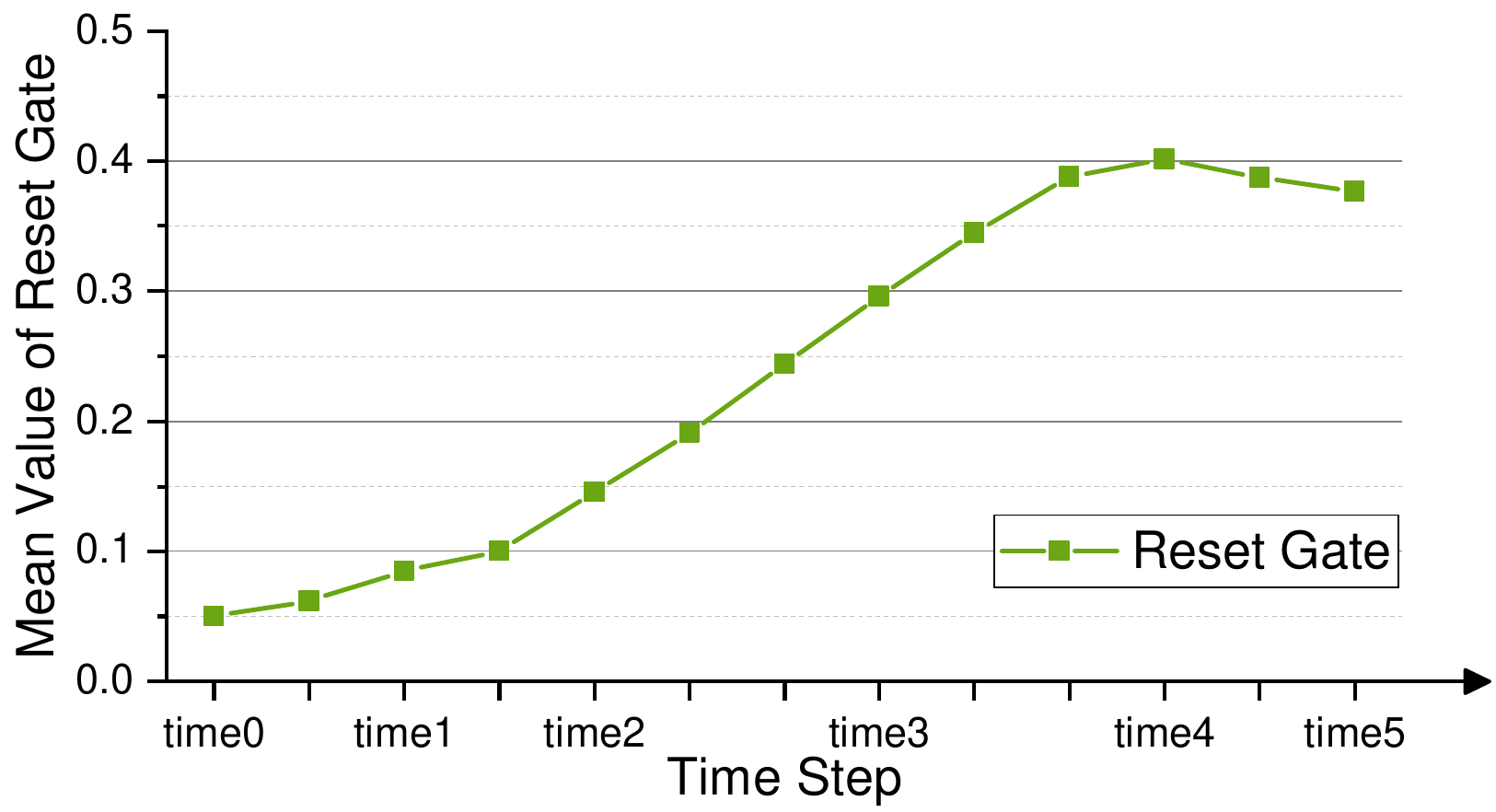} 
		\label{fig:gru_time}
	}
	\caption{An intuitive example to show how HGRN extracts valuable features from neighbors and histories.}.
	\label{fig:intuitive_example}
\end{figure}

To answer \textbf{Q2}, we study the effectiveness of the proposed HGRN network structure, which includes two key components, HGAT and GRU. Ideally, each HGRN agent can not only extract valuable embedding from neighbors but also recall necessary information from its memory, to construct features for decision-making. To provide an intuitive example and get some meaningful insights, we consider a special situation in Surviving.  A screenshot of the considered case is shown in Figure \ref{fig:example}, in which we control agent 0 to pass through agent 1 and agent 2. Then we inspect the graph attention weights, which describes the predicted importance of each agents' information to agent 0, and the mean value of GRU's reset gate, which controls the weight of history features to form the output embedding. The results are shown in Figure \ref{fig:attention_time} and Figure \ref{fig:gru_time}. At time 2, we observe that HGRN manages to pay the most attention to Agent 1 that observes the most food. During time 2 to time 4, in which the HGAT connection is established, the first observation is that the agent pays more attention to itself, possibly since HGRN has well stored the information from its neighbors; meanwhile, the reset gate value of GRU grows rapidly, which denotes that the model relies more on its memory to make the decision. 

\subsubsection{How the self-adaptive temperature works?}
\begin{figure}[t]
	\centering
	\includegraphics[width=0.9\columnwidth]{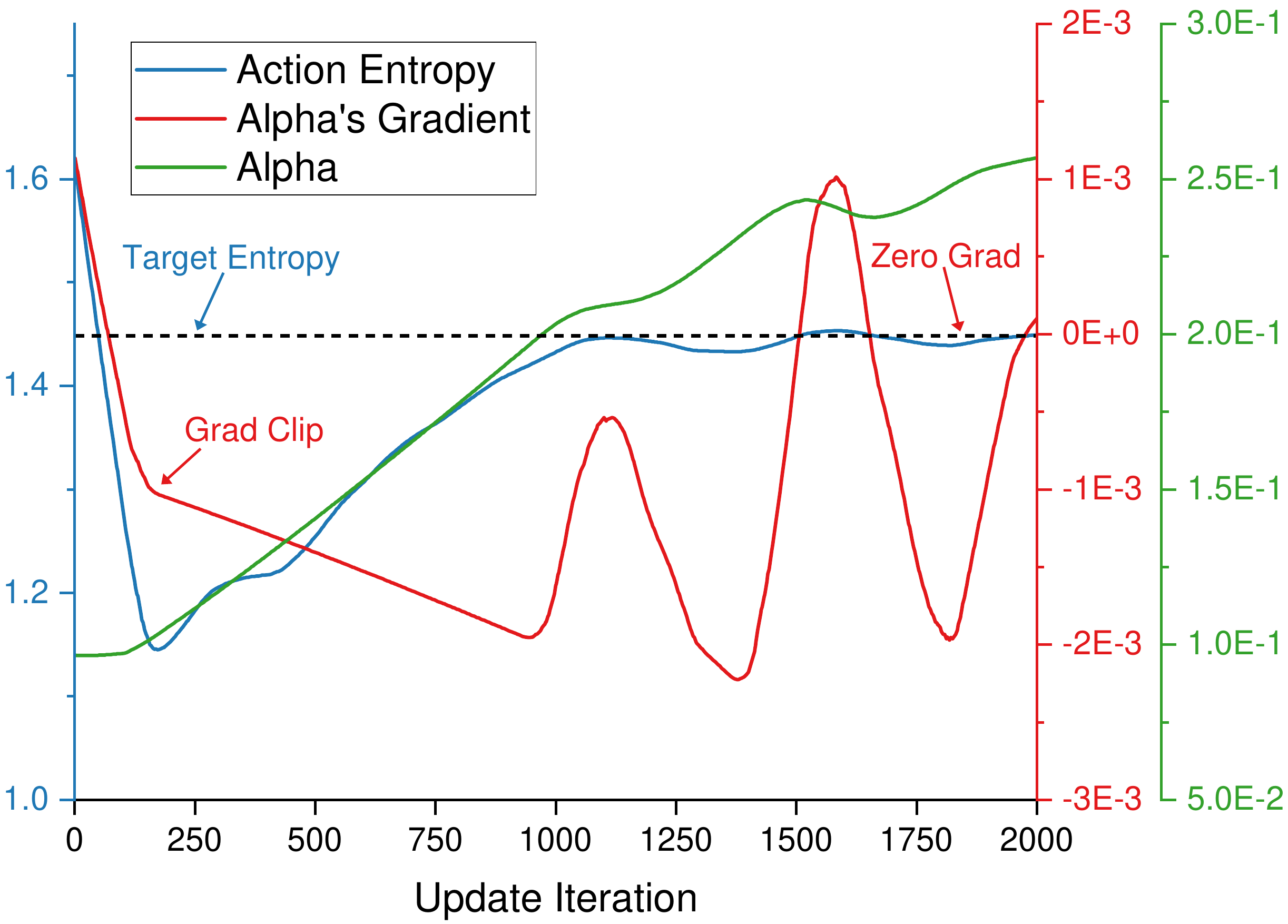} 
	\caption{The learning curves of action entropy, temperature parameter $\alpha$ and its gradient $\nabla\alpha$ during the training process of Soft-HGRN in Surviving. } 
	\label{fig:learn_alpha}
\end{figure}

To answer \textbf{Q3}, we first illustrate the training loop of the learnable temperature parameter $\alpha$. As can be seen in Figure \ref{fig:learn_alpha}, during the training process of Soft-HGRN, when the action entropy is smaller than the target, the model tends to increase $\alpha$ to encourage exploration and vise versa. To prevent instability, we clip the gradient of $\alpha$ when it exceeds the value of $0.01\alpha$, which contributes to the stable convergence of the action entropy. As for the principle behind the better performance of the stochastic policy, our insight is that it enables the agent to learn when/how to explore and exploit. To support our idea, a case study similar to Figure \ref{fig:example} is presented in Appendix. Besides, since the necessity of exploration and the optimal action entropy varies in different environments, we make a parameter analysis on $p_\alpha$ in Appendix.

\subsection{Transferability}
\begin{figure}[t]
	\centering
	\includegraphics[width=0.8\columnwidth]{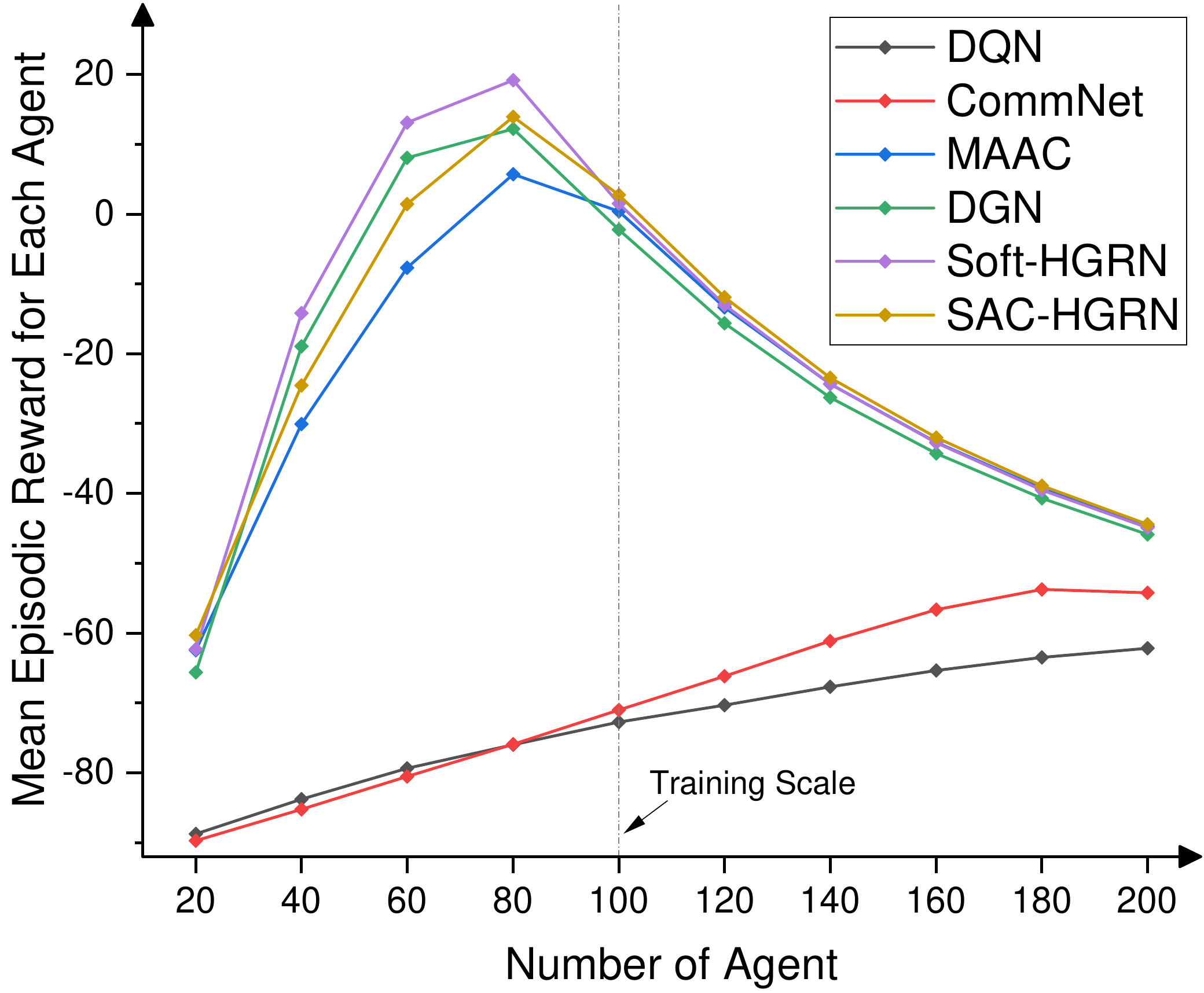} 
	\caption{The comparison of different algorithms in Surviving with various agent scale.} 
	\label{fig:transfer_surviving}
\end{figure}

Weak generalization ability is an urgent problem to be solved by RL algorithms. To test the generalization ability of our model, we train all models in Surviving environment with 100 agents, and then deploy them to environments with different agent scales for testing, as shown in Figure \ref{fig:transfer_surviving}. It can be seen that our approach performs better than the baselines under all tested agent scales. An interesting finding is that Soft-HGRN performs better when the agent scale is less than 100, and SAC-HGRN shows the best performance when the scale is larger than 100. Our insight is that it is easier for Soft-HGRN to estimate the return value of a situation with a small agent scale, yet the discriminative policy of SAC-HGRN is better when the environment becomes too unstable to estimate the return due to the ultra-large agent scale.

\section{Conclusions}
In this paper, we proposed a value-based MADRL model namely Soft-HGRN and its actor-critic variant SAC-HGRN, to address the large-scale multi-agent partially observable problem. The proposed approach consists of two key components, a novel network structure named HGRN that could aggregate information from neighbors and history, as well as a maximum-entropy learning technique that could self-adapt the temperature parameter to learn a stochastic policy with configurable action entropy. The experiment on four many-agent environments demonstrates that the learned model outperforms other MADRL baselines, and ablation studies show the necessity of each component in our approach. We also analyze the interpretability, scalability, and transferability of the learned model. As for future work, we will investigate to transfer our methods into continuous space.

\normalem
\bibliography{aaai22}
\end{document}